\documentclass[10pt,twocolumn,letterpaper]{article}

\usepackage[utf8]{inputenc}
\usepackage[T1]{fontenc}
\usepackage{amsmath,amsthm}
\usepackage{newtxtext,newtxmath}
\usepackage[margin=0.75in,columnsep=0.28in]{geometry}
\usepackage{booktabs}
\usepackage{graphicx}
\graphicspath{{figures/}{./}}
\usepackage{multirow}
\usepackage{enumitem}
\usepackage{hyperref}
\usepackage{xcolor}
\usepackage{float}
\usepackage{caption}
\usepackage{subcaption}
\usepackage{array}
\usepackage{longtable}
\usepackage{tabularx}
\usepackage{url}
\usepackage[numbers]{natbib}

\title{
Governed Shared Memory for Multi-Agent LLM Systems
}

\author{
  Yanki Margalit\textsuperscript{1}\thanks{Corresponding author: \texttt{yanki@caura.ai}},
  Nurit Cohen-Inger\textsuperscript{2},
  Erni Avram\textsuperscript{1},
  Ran Taig\textsuperscript{1},
  and Oded Margalit\textsuperscript{2}, 
  \\[0.5em]
  \textsuperscript{1}Caura.ai \\
  \textsuperscript{2}Computer Science and Information, Ben-Gurion University of the Negev
}

\date{}

\begin{document}

% Span title block + abstract across both columns. Standard idiom for
% twocolumn article docs that don't have a dedicated venue template.
\twocolumn[
  \begin{@twocolumnfalse}
  \maketitle
  \begin{abstract}
  AI memory systems are increasingly moving beyond isolated chatbot histories toward shared state used by fleets of cooperating agents. In this setting, memory is not only a retrieval problem: it becomes a governed distributed-systems problem involving scoped access, temporal correctness, provenance, synchronization, and policy-controlled propagation. We introduce a systems architecture for governed shared memory in multi-agent LLM environments. The architecture formalizes the fleet-memory problem, identifies four failure modes---unauthorized leakage, stale propagation, contradiction persistence, and provenance collapse---and defines primitives for scoped retrieval, temporal supersession, provenance tracking, and policy-governed memory propagation.

  We instantiate these primitives in MemClaw, a production multi-tenant memory service, and evaluate them using ArgusFleet, a reproducible harness that exercises the live REST API along four governance dimensions. Our evaluation is a measurement of one production service rather than a comparison against baselines, and its negative results are central to the contribution. Provenance is the cleanest positive result: all 50 depth-four derivation chains reconstruct with correct writer identity at sub-second per-hop latency. Propagation is correct where measured, with high intra-fleet visibility and no observed cross-fleet leakage; tight per-fact polling shows that under strong write mode, write-to-visible latency is effectively one search round-trip rather than the tens-of-seconds delay suggested by a naive batched probe schedule.

  The live evaluation also surfaces two production-relevant architectural issues. First, scope enforcement was initially asymmetric across API paths: tenant isolation held, but sub-tenant scope was not enforced on direct GET-by-id for agent-scoped credentials; the gap was disclosed and remediated during the study. Second, contradiction supersession works when conflicting writes are admitted, but a synchronous near-duplicate gate can reject contradictory writes before the asynchronous contradiction detector observes them. These findings support the central claim that long-context retrieval alone is insufficient for production multi-agent memory: governed shared memory requires explicit systems-level abstractions, and live evaluation is necessary to expose the enforcement and pipeline-ordering failures that design-only treatments miss.
  \end{abstract}
  \vspace{1em}
  \end{@twocolumnfalse}
]

\section{Introduction}

\begin{figure*}[t!]
\centering
\includegraphics[width=0.85\linewidth]{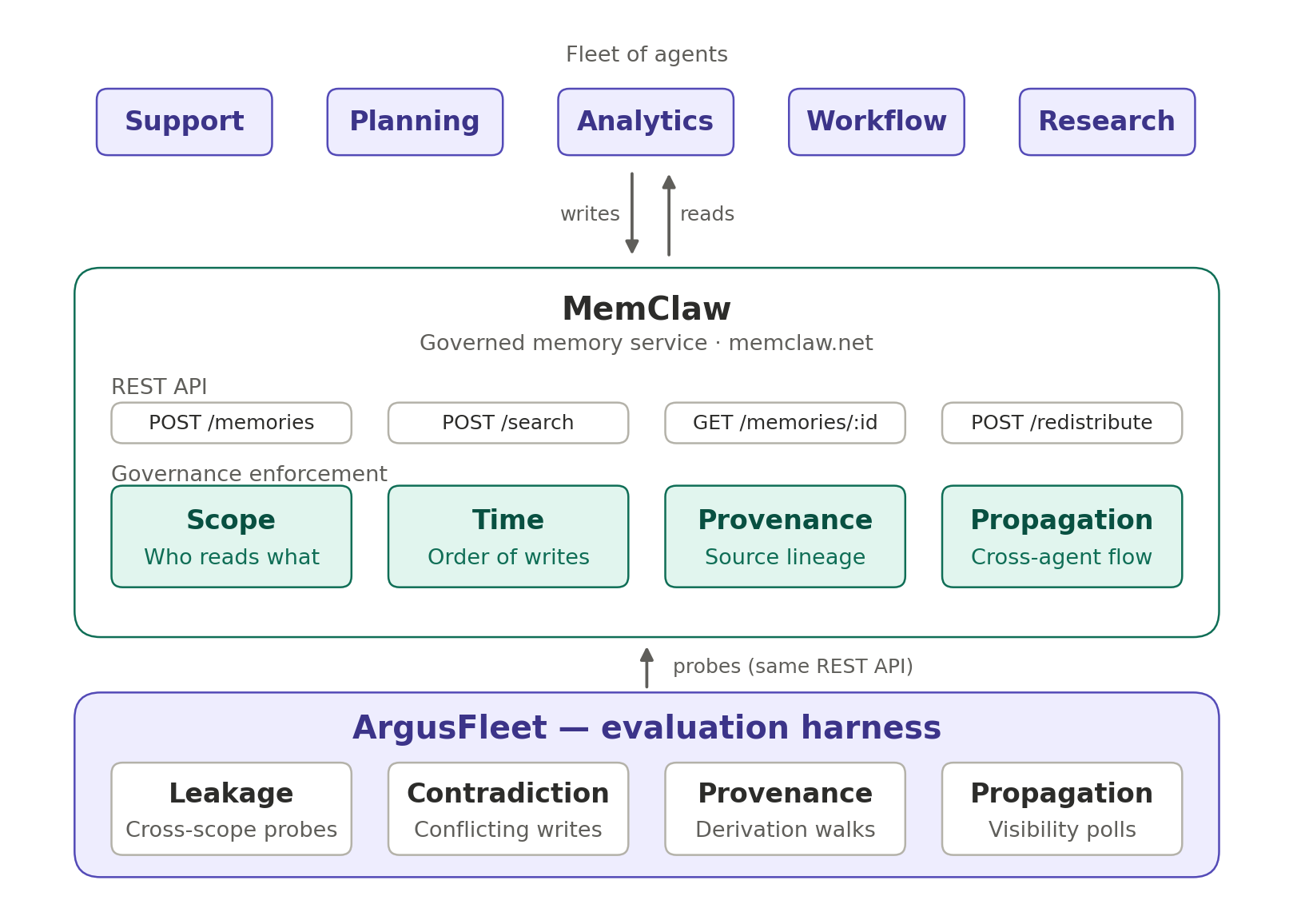}
\caption{System overview. A fleet of cooperating agents writes to
and reads from governed shared memory through MemClaw's REST API. MemClaw
exposes four governance dimensions---scope, time, provenance, and
propagation---corresponding to who may read a memory, which version is
current, where a memory came from, and how it moves across agent boundaries.
ArgusFleet probes the same API surface with one experiment per dimension:
leakage probes scope enforcement, contradiction probes temporal supersession,
provenance walks derivation chains, and propagation measures authorized
visibility and cross-fleet leakage. Stale propagation is evaluated jointly
through contradiction supersession, where outdated rows are marked non-active,
and propagation latency, where newly written facts are polled until visible to
authorized readers.}
\label{fig:overview}
\end{figure*}

Large-language-model systems are rapidly evolving from isolated assistants into fleets of interacting agents. Enterprise copilots, workflow orchestrators, customer-support systems, research agents, and automation pipelines increasingly rely on persistent memory to coordinate behavior across tasks and time. Most existing memory systems, however, were designed around a single-agent abstraction --- one user, one conversation, one retrieval stream, one context window --- and this assumption breaks under fleet-scale deployments.

When multiple agents interact through shared memory, new correctness and governance requirements emerge that are absent from the single-agent setting. The system must answer who is allowed to retrieve which memory, how conflicting facts are resolved when two agents disagree, how knowledge propagates safely across agent boundaries, whether each retrieved memory can be traced back to its writer, and how stale memories are invalidated when the underlying state changes. These are not retrieval-quality questions. They are distributed-systems and database-consistency questions in retrieval clothing.

Current AI memory systems largely optimize for properties inherited from the single-agent world: retrieval relevance, conversational recall, long-context compression, and semantic similarity. Production multi-agent deployments require an additional layer on top: scoped visibility, policy enforcement, temporal correctness, synchronization semantics, and provenance guarantees. We argue that AI memory is therefore evolving from a context-window problem into a governed distributed-memory problem, and that this transition warrants the same systems-level attention historically devoted to databases, event-sourcing, and synchronization infrastructure.

This paper introduces a systems architecture for governed shared memory under that framing and presents MemClaw as a reference implementation deployed at fleet scale. We also introduce ArgusFleet, an evaluation harness that lets the resulting primitives be measured empirically against a live production service.

\subsection{Contributions}

This paper makes three contributions.

First, we formalize the fleet-memory problem for multi-agent LLM systems, modeling shared memory as governed operational state with scope, provenance, and temporal supersession. We identify four resulting failure modes: unauthorized leakage, stale propagation, contradiction persistence, and provenance collapse.

Second, we propose an architecture for governed shared memory, centered on scoped retrieval, temporal contradiction resolution, provenance tracking, and policy-governed propagation. We instantiate the architecture in MemClaw, a production multi-tenant memory service, and introduce ArgusFleet, a reproducible harness that probes these primitives through the live REST API.

Third, we report a live-service measurement of MemClaw using ArgusFleet. The results show complete provenance reconstruction and correct propagation where measured, but also expose two production-relevant issues: a remediated \(\mathtt{GET}\)-by-id sub-tenant scope gap and a pipeline-ordering interaction in which synchronous deduplication can pre-empt asynchronous contradiction detection.

\section{Background and Related Work}

\subsection{Long-Context Retrieval}

Long-context and retrieval-augmented methods improve how much relevant history a model can access~\citep{longcontext_survey,rag2020}. But they primarily optimize recall within an isolated interaction, not governance. Larger context windows do not by themselves provide scoped access, temporal consistency, provenance, or synchronization; multi-agent deployments therefore require memory systems that answer not only \emph{can the model retrieve this}, but \emph{should this agent see this version}.

\subsection{Agent Memory Systems}

A line of recent agent-memory architectures --- including MemGPT and its successor Letta~\citep{memgpt2023}, Mem0~\citep{mem0}, Zep~\citep{zep}, LangMem~\citep{langmem}, A-MEM~\citep{amem2025}, and the memory subsystems of AutoGen~\citep{autogen} --- treats memory as a stand-alone component layered on top of an LLM. Most assume single-agent retrieval against an append-only store with largely unconstrained access, weak provenance semantics, and limited temporal resolution. These assumptions are reasonable for chatbot-style interactions in which one user converses with one assistant; they become problematic under multi-agent shared-memory workloads where different agents have different visibility, different write authority, and different reasoning windows.

A second, very recent line moves explicitly to \emph{shared} multi-agent memory and is the closest neighbour to our setting. G-Memory~\citep{gmemory2025} formalizes multi-agent memory as a three-tier interaction/query/insight graph and argues, as we do, that single-agent memory does not transfer to multi-agent systems; MIRIX~\citep{mirix2025} organises a multi-agent memory across typed stores; and Collaborative Memory~\citep{collabmemory2025} adds \emph{dynamic access control} and immutable provenance over a shared multi-user store. These systems establish that shared memory, governance, and provenance are an active concern --- but none treats temporal contradiction-resolution and supersession as a first-class memory operation, and none reports enforcement measured against a live service. Our contribution is the unification of all five components (\(A, M, G, P, T\)) under one formalism, with temporal supersession (\(T\)) as the element prior shared-memory work leaves implicit, and an empirical harness that exercises each against a running system.

\subsection{Distributed Systems and Shared State}

Once memory is genuinely shared across agents, the operational concerns mirror those of traditional distributed systems: synchronization, consistency, access governance, temporal ordering, event propagation, and auditability. Production AI memory systems therefore increasingly resemble distributed databases~\citep{dynamo2007, spanner2013}, event-sourcing and weakly-connected replicated stores~\citep{bayou1995, crdt2011}, synchronization layers~\citep{lamport1978}, and governance middleware~\citep{rbac1996, nist_attribute_based} --- not the retrieval indexes from which the agent-memory line of work descended. This suggests that memory infrastructure may become a foundational systems layer for future AI architectures, and that the methods and primitives developed for distributed state are the right vocabulary for reasoning about it.

\subsection{Memory Governance, Security, and Privacy}

A parallel and equally recent line treats agent memory as a security and privacy surface rather than a retrieval index. MEXTRA~\citep{mextra2025} shows that an agent's memory module is a concretely extractable attack surface: crafted queries alone recover private user--agent interactions in a black-box setting, motivating exactly the scoped-retrieval guarantee we formalize. On the enforcement side, CaMeL~\citep{camel2025} attaches capability and provenance metadata to values and enforces policy at tool-call time, while Fides~\citep{fides2025} gives a formal information-flow-control model with confidentiality and integrity labels and a planner that deterministically enforces them. For the propagation question --- whether a memory \emph{should} cross an agent boundary --- the canonical framing is contextual integrity: ConfAIde~\citep{confaide2024} shows that capable LLMs leak information in contexts humans would withhold, reframing access as a question of appropriate flow rather than static permission. Our unauthorized-leakage finding (\S\ref{sec:findings-bimodal}) is an instance of the classic confused-deputy problem this literature studies: a read handler that resolves but then ignores caller identity. We adopt these as the governance and provenance vocabulary for \(G\) and \(P\), complementing the distributed-systems vocabulary above for \(M\) and \(T\).

Finally, our evaluation framing is also related to recent work that treats LLM evaluation itself as a multi-agent process. PeerRank evaluates models through web-grounded, bias-controlled peer review among LLMs~\citep{margalit2026peerrankautonomousllmevaluation}. Our use of ArgusFleet is complementary: rather than evaluating model outputs, it evaluates the governed memory substrate that multi-agent systems rely on for persistent shared state.

\section{The Fleet-Memory Problem}
\label{sec:problem}

Modern AI deployments are increasingly composed not of a single assistant but of fleets of specialised agents operating over shared persistent state. A customer-support agent may update a user's billing record in a way that subsequently affects a planning agent's schedule, an analytics agent's dashboard, a workflow orchestrator's routing logic, and a downstream compliance system's audit trail. A research agent may surface a finding that synthesis agents, verification agents, reporting systems, and autonomous workflows all need to consume hours or days later. In these environments, memory is no longer ``conversation history''; it is shared operational state, and the correctness story it must support has more in common with shared-storage subsystems than with retrieval indexes.

This shift changes the nature of the problem. Traditional LLM memory architectures optimize for semantic recall, conversational continuity, long-context retrieval, and summarisation quality. These optimizations are sufficient when one agent interacts with one user over one conversational stream. They break down when many agents write to shared memory, when visibility differs across roles, when facts evolve over time, and when correctness matters operationally rather than only narratively. Under fleet-scale workloads, the memory subsystem has to answer which agents are allowed to retrieve which memories, what happens when two agents write contradictory facts, how stale memories are superseded, whether every retrieved memory can be traced to its source, and how knowledge propagates safely across agent boundaries. These are distributed-systems questions, not retrieval questions.

\begin{quote}
\emph{AI memory is evolving from a context-window problem into a distributed systems problem.}
\end{quote}

The transition introduces a new set of requirements --- governance, temporal correctness, synchronization, provenance, scoped visibility, and policy-aware retrieval --- that the single-agent memory line of work does not address. We refer to systems that meet these requirements as \textbf{fleet-memory systems}.

\subsection{Fleet-Memory Systems}

We define a fleet-memory system as:

\[
F = (A, M, G, P, T)
\]

where:
\begin{itemize}
    \item \(A\) represents a set of interacting agents,
    \item \(M\) is a shared memory substrate,
    \item \(G\) is a governance and policy layer,
    \item \(P\) represents provenance metadata,
    \item \(T\) defines temporal ordering and supersession semantics.
\end{itemize}

Unlike single-agent conversational memory, fleet-memory systems must maintain correctness across interacting reads and writes performed by multiple autonomous actors over time.

This introduces challenges analogous to:
\begin{itemize}
    \item distributed consistency,
    \item access control,
    \item synchronization,
    \item event sourcing,
    \item and state resolution.
\end{itemize}

The central challenge is no longer merely retrieving semantically relevant information; it is maintaining operationally correct shared state.

\subsection{Memory Operations}

A memory write operation is defined as:

\[
w_i = (a_i, c_i, s_i, t_i, p_i)
\]

where:
\begin{itemize}
    \item \(a_i\) is the writing agent,
    \item \(c_i\) is the memory content,
    \item \(s_i\) defines the visibility scope,
    \item \(t_i\) is the temporal ordering,
    \item \(p_i\) contains provenance metadata.
\end{itemize}

Importantly, writes are not immutable conversational artifacts but state transitions: a later write may supersede, invalidate, restrict, or contradict prior memory. Retrieval operations must therefore incorporate governance and temporal semantics in addition to semantic relevance.

We define retrieval as:

\[
r_j = f(q_j, a_j, g_j, t_j)
\]

where:
\begin{itemize}
    \item \(q_j\) is the retrieval query,
    \item \(a_j\) is the requesting agent,
    \item \(g_j\) represents governance constraints,
    \item \(t_j\) defines temporal correctness conditions.
\end{itemize}

In conventional vector retrieval systems, semantic similarity alone determines retrieval eligibility; in fleet-memory systems, retrieval correctness additionally depends on:
\begin{itemize}
    \item authorization,
    \item state freshness,
    \item contradiction resolution,
    \item provenance validity,
    \item and synchronization guarantees.
\end{itemize}

We make the authorization requirement precise, since it is the one the
evaluation falsifies. Let \(\mathrm{auth}(a_j, s_i, G)\) denote the predicate
``agent \(a_j\) is entitled, under governance \(G\), to view a memory carrying
scope \(s_i\),'' where \(s_i\) decomposes into nested levels (agent
\(\sqsubseteq\) fleet \(\sqsubseteq\) tenant; \S\ref{sec:impl}). A retrieval is
\emph{scope-sound} when
\[
w_i \in r_j \;\Longrightarrow\; \mathrm{auth}(a_j, s_i, G).
\tag{\textsc{Inv-Scope}}
\]
That is, no agent receives a memory its identity is not entitled to under the
row's scope. \textsc{Inv-Scope} is the formal content of the unauthorized-leakage
failure mode (\S\ref{sec:failures}): a leak is exactly a violation of it. A
correct governed-memory implementation must evaluate the \emph{full} scope
predicate on every retrieval path; as \S\ref{sec:results} shows, the
\(\mathtt{GET}\)-by-id path evaluated only the tenant projection of \(s_i\) at
measurement time and so violated \textsc{Inv-Scope} for any fleet- or
agent-scoped row. The formalism predicted a concrete violation, the experiment
measured it, and the operator's fix (\S\ref{sec:findings-bimodal}) restored
\textsc{Inv-Scope} on that path --- a predict\(\to\)measure\(\to\)remediate loop.

This distinction becomes critical once memory evolves from passive context into shared operational infrastructure.

\section{Fleet-Memory Failure Modes}
\label{sec:failures}

We identify four primary fleet-memory failure classes.

\subsection{Unauthorized Leakage}

An agent retrieves memory outside its authorized scope.

\paragraph{Example}
A customer-support agent retrieves billing notes intended only for finance agents.

\paragraph{Implications}
Unauthorized leakage introduces privacy risk, tenant-isolation failures, compliance exposure, and operational unpredictability. Naive semantic retrieval architectures are especially vulnerable because retrieval eligibility is governed primarily by embedding similarity rather than by explicit policy enforcement.

\subsection{Stale Propagation}

Memory updates fail to synchronize correctly across agents.

\paragraph{Example}
One agent updates a user's shipping address while another agent continues retrieving outdated state.

\paragraph{Implications}
Stale propagation produces inconsistent workflows, contradictory actions, degraded user trust, and operational instability. The eventual-consistency window between write and read becomes part of the system's latency budget, not an implementation detail.

\subsection{Contradiction Persistence}

Conflicting memories coexist without resolution.

\paragraph{Example}
Two mutually incompatible user preferences remain simultaneously retrievable.

\paragraph{Implications}
Append-only retrieval systems often lack explicit supersession semantics, allowing outdated or contradictory memories to persist indefinitely. Downstream agents that read both have no principled way to choose between them.

\subsection{Provenance Collapse}

Retrieved memories cannot be traced to their origin.

\paragraph{Example}
An agent retrieves a fact without attribution to the original writer, source system, or timestamp.

\paragraph{Implications}
Without provenance, debugging becomes guesswork, auditability weakens, and governance guarantees become unverifiable: the system cannot demonstrate \emph{how} a particular fact entered the store, only that it is present.

\section{Governed Shared Memory Architecture}

\subsection{Design Principles}

The proposed architecture is built around five principles:
\begin{enumerate}
    \item Scoped retrieval
    \item Explicit provenance
    \item Temporal correctness
    \item Policy-governed propagation
    \item Persistent shared state
\end{enumerate}

\subsection{Memory Scopes}

We define four memory scopes:

\begin{table}[H]
\centering
\caption{Memory scopes}
\begin{tabular}{ll}
\toprule
Scope & Description \\
\midrule
Agent-local & Visible only to one agent \\
Team-shared & Shared among a defined agent group \\
Tenant-global & Shared across a tenant environment \\
Restricted & Explicitly policy-constrained \\
\bottomrule
\end{tabular}
\end{table}

Semantic similarity alone is insufficient for retrieval eligibility; every retrieval operation must additionally satisfy the governance policies attached to the scope of the candidate row.

\subsection{Temporal Resolution}

The architecture treats memory as evolving state rather than immutable append-only retrieval. Each memory object carries a creation timestamp, optional supersession references, contradiction markers, provenance metadata, and a confidence state, and contradictory writes are resolved through temporal ordering and policy-aware resolution rules rather than by silently coexisting in the store.

\subsection{Provenance Graph}

Every memory object stores its writer identity, source system, derivation history, and modification lineage. This metadata is what enables auditability, debugging, traceability, and compliance verification --- four properties that are increasingly required of memory systems deployed in regulated environments and that distinguish governed memory from retrieval indexes.

\subsection{Policy-Governed Retrieval}

Retrieval consists of:
\begin{enumerate}
    \item semantic candidate generation,
    \item policy filtering,
    \item temporal resolution,
    \item provenance enrichment,
    \item ranked delivery.
\end{enumerate}

This differs from traditional vector retrieval systems in which semantic similarity alone determines retrieval eligibility.

\section{Reference Implementation}
\label{sec:impl}

We instantiate the architecture of \S5 in two artifacts: MemClaw, a multi-tenant memory service, and ArgusFleet, an open evaluation harness that exercises the architectural primitives against MemClaw or any wire-compatible implementation.

\subsection{MemClaw}

MemClaw is a multi-tenant governed-memory service\footnote{\url{https://memclaw.net}}. Each memory object carries the fields required by the architecture: tenant and fleet identifiers, writing agent, content, optional RDF triple (\(\langle\) subject\_entity, predicate, object\_value \(\rangle\)), visibility scope, status, and supersession links. The wire format additionally exposes ingestion-side metadata (write mode, enrichment-pending flags) so a caller can reason about consistency at write time. Authentication uses an \texttt{mc\_}-prefixed API key; the credential's \emph{kind} (tenant-scoped vs.\ agent-scoped) is bound on the credential record, not encoded in the prefix. For an agent-scoped credential the gateway resolves the caller's agent identity and injects it (as \texttt{X-Agent-ID}) before request handlers execute --- a client-supplied \texttt{X-Agent-ID} is stripped --- while tenant boundaries are enforced on every path.

The following operations are exercised in this paper:
\begin{itemize}
    \item \texttt{POST /api/v1/memories} --- write a memory with optional RDF triple and write mode (\texttt{fast}, \texttt{strong}, \texttt{auto}, or \texttt{stm}).
    \item \texttt{POST /api/v1/search} --- semantic search filtered by tenant and optional fleet/agent/status constraints; top-\(k\) is clamped to \(20\) on the REST surface.
    \item \texttt{POST /api/v1/entities/upsert} --- materialize a subject entity for RDF-triple writes.
    \item \texttt{GET /api/v1/memories/\{id\}} --- retrieve a memory by identifier; requires the tenant identifier as a query parameter.
    \item \texttt{POST /api/v1/memories/redistribute} --- re-home a memory onto a different agent; requires the caller to hold \(\text{trust\_level} \ge 3\).
\end{itemize}

\subsection{ArgusFleet}

ArgusFleet is an open-source Python 3.12 evaluation harness\footnote{\url{https://github.com/caura-ai/argusfleet}} for governed shared memory, comprising four experiments, an async REST client, a typed domain model, structured event logging, and a reproducible reporting pipeline. It is distributed as the \texttt{argusfleet} package; we use the name ArgusFleet for the system throughout this paper. The harness produces JSONL event traces, CSV summaries, and Matplotlib figures from each run, all derived from the same canonical event stream so analysis can be re-executed without re-hitting the server. The event traces backing every number in this paper --- including all four \(N{=}200\) contradiction runs and the \(N{=}100\) comparison (\S\ref{sec:results}) --- are committed under \texttt{traces/} in that repository (commit \texttt{2c55bb5}); each experiment's reported figures and \texttt{summary.csv} regenerate deterministically from its per-write \texttt{events.jsonl}.

\paragraph{Determinism.}
Every workload is seeded. Plan structure (fleets, agents, fact subjects, chain shapes) is reproducible across runs sharing the same seed. Per-run content is salted with a run-unique nonce so that re-running against a stateful production service still produces semantically fresh writes rather than collisions on accumulated state. Each run materializes a uniquely-named output directory of the form \texttt{<UTC-timestamp>-<experiment>-seed<N>/}.

\paragraph{Async execution.}
The client is built on \texttt{httpx} and \texttt{tenacity} with exponential backoff for transient transport errors, 5xx responses, and \(429\) rate-limit responses; other 4xx responses are surfaced to the experiment unmodified, since they are precisely the governance signal we are measuring. Concurrency caps are configured per experiment to remain below the per-tenant search rate limit (\(100\) req/min on \texttt{/search}).

\paragraph{Domain model.}
Per-probe and per-write events are appended to the JSONL trace as
plain dictionaries; the harness ships typed Pydantic models for the two
records where typing actually pays off --- \texttt{MemoryScope} (the
ground-truth scope enum the leakage experiment reasons about) and
\texttt{PolicyDecision} (the audit-grade record of an observed
governance outcome). Translating between the dict events and the
MemClaw wire schema is delegated to the async client (\S6.1) so the
harness can be retargeted to alternative governed-memory implementations
without changing the experiment code.

\section{Evaluation Methodology}
\label{sec:method}

We evaluate the four failure modes from \S4 with one experiment each. Every experiment shares the same lifecycle: a deterministic plan is materialized from the seed, writes are issued against the live service, a settle delay accommodates asynchronous server-side enrichment, and a verification phase compares observed behaviour against an expected ground truth derived from the plan.

\subsection{Leakage Experiment}

We instantiate \(F = 4\) fleets, each containing \(K = 6\) agents, in a single tenant. The first agent in each fleet (the \emph{owner}) writes two secrets: one with fleet scope, one with agent scope, each containing a unique canary token in metadata and substantively distinct prose in content. Every agent then probes every secret by issuing a search whose query is the secret's prose; the harness classifies each probe against ground truth (does this agent's tenant/fleet/agent identity entitle it to see this secret?) and labels the result.

For each probe, we record two orthogonal failure signals:
\begin{itemize}
    \item \textbf{leak} --- the server returned a row the probe was not entitled to see; this is the security failure that motivates governance.
    \item \textbf{miss} --- the server failed to return a row the probe was entitled to see; this is an availability failure that reflects retrieval recall.
\end{itemize}

We report \(\textsc{leak\_rate} = n_{\text{leaks}} / n_{\text{expected-deny}}\) and \(\textsc{miss\_rate} = n_{\text{misses}} / n_{\text{expected-allow}}\). At the configured workload size we generate \(192\) probes spanning \(164\) expected-deny and \(28\) expected-allow.

To separate an authentication-model effect from a genuine enforcement gap on the access path, we additionally run a focused \emph{agent-scoped} probe. Using the tenant key we self-provision two agent credentials (\texttt{POST /api/v1/admin/agent-keys/provision}, \(\mathtt{kind{=}agent\_key}\)) bound to the same fleet but different trust levels --- one at \(\text{trust}=1\) (write home fleet; below the cross-fleet-read rung) and one at \(\text{trust}=2\) (cross-fleet read). We confirm the gateway resolves each agent's identity via \texttt{GET /api/v1/whoami}, then \(\mathtt{GET}\) a row written under a \emph{different} fleet with each credential. Under the documented trust ladder (\(0\) read-only, \(1\) write-home-fleet, \(2\) cross-fleet-read, \(3\) cross-fleet-write), the trust-\(1\) cross-fleet read must be denied if sub-tenant scope is enforced on \(\mathtt{GET}\).

\subsection{Contradiction Experiment}

We construct \(100\) facts drawn from \(20\) distinct subject--predicate templates (e.g., \(\langle \text{billing-service-N}, \text{deployment\_region} \rangle\)). For each fact we run two scenarios: a \emph{sequential} writer pair separated by a small delay, and a \emph{concurrent} pair dispatched simultaneously. Within each pair, both writes share the same \texttt{subject\_entity\_id} and \texttt{predicate} but assert different \texttt{object\_value}. The RDF triple is what we expect to route the second write into the server's structural contradiction detector rather than into the embedding-similarity dedup gate.

We report:
\begin{itemize}
    \item \textbf{detection\_rate} --- the fraction of fact-runs in which, after a settle window, a \(\mathtt{GET}\)-by-id \emph{re-fetch} of the two rows shows a supersession edge (\texttt{supersedes\_id}) or the older row flipped to a non-active status. The re-fetch is essential: contradiction detection runs \emph{post-commit and asynchronously}, so \texttt{supersedes\_id} is never present in the synchronous write response (reading it there --- as an earlier version did --- yields a spurious \(0\)). The predicate must also be one the server treats as single-valued, or its structural detector is skipped.
    \item \textbf{stale\_read\_rate} --- the fraction of fact-runs in which a fresh \texttt{status\_filter="active"} search returned more than one row carrying the fact identifier.
    \item \textbf{write\_latency} percentiles, split by whether the write triggered a contradiction.
\end{itemize}

\subsection{Provenance Experiment}

We construct \(50\) derivation chains of depth \(4\) (\(200\) total writes). Each chain corresponds to a synthetic incident narrative drawn from a pool of \(25\) operational topics (cycled, with a per-chain nonce keeping content distinct) (e.g., \emph{checkout error rate spike}, \emph{Kafka consumer rebalance loop}). The four steps in each chain correspond to \emph{observation}, \emph{hypothesis}, \emph{mitigation}, and \emph{verification}. Each step writes a memory with \texttt{metadata.derived\_from} encoding the parent identifier. After all writes settle, the harness fetches the leaf and walks the chain back to the root, grading three signals:
\begin{itemize}
    \item \textbf{completeness} --- every ancestor on the planned chain is reachable via \texttt{GET}.
    \item \textbf{accuracy} --- at each step the observed writer matches the planned writer.
    \item \textbf{depth\_fidelity} --- the reconstructed chain length equals the planned depth.
\end{itemize}

\subsection{Propagation Experiment}

The experiment has two deliberately separate phases, because the visibility \emph{rate} and the consistency \emph{latency} want opposite measurement regimes.

\paragraph{Rate phase.} A designated writer agent writes \(40\) fleet-scoped facts, each phrased as a substantively distinct operational statement (drawn from \(40\) templates). \(3\) sibling readers in the same fleet and \(2\) non-fleet readers in the same tenant then poll \texttt{/search} for the prose (\(200\) probes total). Fleet-sibling probes time out after \(30\) seconds; non-fleet probes execute a single attempt and any hit is counted as a cross-fleet leak. This phase scales with the fact count to tighten the rate CIs; its per-probe latency is \emph{not} used (see below).

\paragraph{Window phase.} Because the rate phase seeds every fact before draining any probe through a concurrency gate, a probe's first \texttt{/search} lands long after the row is already searchable --- so its time-to-visibility measures harness scheduling, not server lag, and inflates with probe count. We therefore measure the write-to-visible window separately, on \(8\) dedicated facts (distinct prose, so they do not dedup-collide with the rate-phase set), each polled \emph{immediately and continuously} from write completion at \(250\) ms granularity, one fact at a time so the poll stream stays under the per-tenant \texttt{/search} budget. This isolates the server's single-write consistency window from harness scheduling.

For each (writer, fact, reader) tuple we record:
\begin{itemize}
    \item \textbf{visibility} --- whether the reader's search ever returned an item carrying the fact's canary in metadata (rate phase).
    \item \textbf{time-to-visibility} --- elapsed wall-clock between write completion and the first successful read, from the window phase.
    \item \textbf{is\_leak} --- a foreign-fleet reader observed the canary.
\end{itemize}

A deterministic subset of facts is additionally passed through \texttt{POST /api/v1/memories/redistribute} onto a target agent, and the harness re-fetches each row via \texttt{GET} to confirm that \texttt{agent\_id} flipped to the target.

\section{Experimental Results}
\label{sec:results}

We ran the four experiments against the production \texttt{memclaw.net} service from a single tenant. Results are summarised in Table~\ref{tab:headline}.

\paragraph{Temporal provenance.} All four experiments report a freshly-provisioned tenant measured on 2026-05-30. One finding --- sub-tenant scope on the \(\mathtt{GET}\)-by-id path (\S\ref{sec:findings-bimodal}) --- was disclosed and remediated server-side on 2026-05-31; we report both the as-measured state and the verified fix, and flag the affected numbers as ``as measured'' where they appear. Every other number reflects the 2026-05-30 state, and all are backed by committed traces (\S\ref{sec:impl}). Tenses in the body are relative to these two dates.

\begin{table*}[t]
\centering
\small
\caption{Headline results across the four experiments at \(N = 200\) (planned probes/writes per experiment) on a freshly-provisioned MemClaw tenant. Leakage carries two signals per probe: the authoritative \emph{access} signal from \(\mathtt{GET}\)-by-id, and the secondary \emph{discovery} signal from semantic search. Counts in parentheses indicate the denominator the rate is over. Each rate is followed by its \(95\%\) Wilson score interval. The GET-by-id tenant-key rows measure tenant-wide reachability under a tenant-scoped credential and should be read as exposure under tenant authority, not by themselves as a sub-tenant policy violation. The sub-tenant enforcement gap is established by the agent-scoped probe in Table~\ref{tab:agentscope}: at measurement time, a trust-\(1\) agent could \(\mathtt{GET}\) a cross-fleet row that the trust ladder should have denied; this was remediated on 2026-05-31 (\S\ref{sec:findings-bimodal}; re-probe trust-\(1\) cross-fleet \(0/36\)).}
\label{tab:headline}
\begin{tabularx}{\linewidth}{lXr}
\toprule
Experiment & Metric & Value \\
\midrule
\multirow{10}{*}{Leakage}
  & probes total (planned / executed) & \(192\) / \(192\) \\
  & tenant-key GET exposure, GET-by-id (\(n_{\text{cross-scope}}=164\)) & \(\mathbf{1.000}\) \\
  & \quad 95\% Wilson CI & \([0.977, 1.000]\) \\
  & tenant-key GET miss rate, GET-by-id (\(n_{\text{in-scope}}=28\)) & \(0.000\) \\
  & \quad 95\% Wilson CI & \([0.000, 0.121]\) \\
  & \textbf{search} leak rate, semantic search (\(n_{\text{deny}}=164\)) & \(0.439\) \\
  & \quad 95\% Wilson CI & \([0.365, 0.516]\) \\
  & \textbf{search} miss rate (\(n_{\text{allow}}=28\)) & \(0.143\) \\
  & \quad 95\% Wilson CI & \([0.057, 0.315]\) \\
  & access / search latency p50 & \(296\) / \(726\) ms \\
\midrule
\multirow{5}{*}{Provenance}
  & chain completeness (depth \(=4\), \(50\) chains) & \(\mathbf{1.000}\) \\
  & \quad 95\% Wilson CI & \([0.929, 1.000]\) \\
  & writer-identity accuracy (all chains) & \(1.000\) \\
  & \quad 95\% Wilson CI & \([0.929, 1.000]\) \\
  & per-hop fetch latency p50 / p95 / p99 & \(291\) / \(491\) / \(1{,}076\) ms \\
\midrule
\multirow{7}{*}{Propagation}
  & writes succeeded (planned \(=40\)) & \(40\) \\
  & visibility probes produced (planned \(=200\)) & \(200\) \\
  & fleet-sibling visibility (\(n_{\text{probes}}=120\)) & \(\mathbf{0.975}\) \\
  & \quad 95\% Wilson CI & \([0.929, 0.991]\) \\
  & cross-fleet leak rate (\(n_{\text{probes}}=80\)) & \(\mathbf{0.000}\) \\
  & \quad 95\% Wilson CI & \([0.000, 0.046]\) \\
  & write\(\to\)visible window p50 / p95 (tight poll, \(8\) facts) & \(0.83\) / \(1.63\) s \\
\midrule
\multirow{7}{*}{Contradiction}
  & writes succeeded (planned \(=400\)) & \(194\) \\
  & detection rate, all fact-runs (\(n=200\)) & \(0.490\) \\
  & \quad 95\% Wilson CI & \([0.422, 0.559]\) \\
  & detection rate, both writes admitted (\(n=90\)) & \(\mathbf{1.000}\) \\
  & \quad 95\% Wilson CI & \([0.959, 1.000]\) \\
  & write latency p50 / p95 (succeeded) & \(1{,}840\) / \(4{,}861\) ms \\
  & write latency p99 & \(19{,}319\) ms \\
\bottomrule
\end{tabularx}
\end{table*}

\subsection{Leakage}

\begin{figure}[H]
\centering
\includegraphics[width=\linewidth]{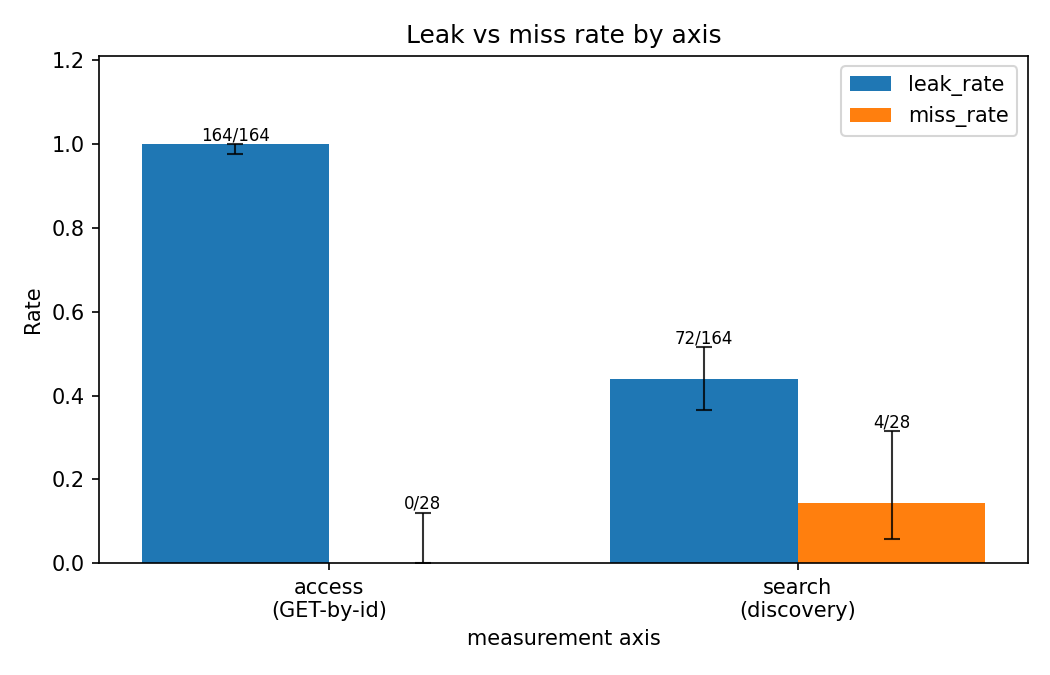}
\caption{Leakage envelope by measurement axis, \emph{as measured} (2026-05-30). The \textsc{leak\_rate} (over expected-deny probes) is the security signal; the \textsc{miss\_rate} (over expected-allow probes) is the availability signal. Enforcement was bimodal. The \(\mathtt{GET}\)-by-id tenant-key axis returned every requested row to a tenant-scoped caller (tenant-key exposure \(=1.000\)), which is expected under tenant-wide authority. The sub-tenant enforcement gap is shown by the agent-scoped probe in Table~\ref{tab:agentscope}: a trust-\(1\) agent could retrieve a cross-fleet row that should have been denied. The discovery path applied the fleet filter only partially (\textsc{search\_leak\_rate} \(=0.439\)) and missed a seventh of entitled probes under embedding-similarity ranking.}
\label{fig:leakage}
\end{figure}

Across \(192\) probes spanning \(164\) expected-deny and \(28\) expected-allow combinations, the two-signal measurement (\S\ref{sec:method}) returns markedly different rates on the two axes; see Fig.~\ref{fig:leakage}. The headline finding is that \emph{scope enforcement was bimodal at measurement time} (the \(\mathtt{GET}\)-path gap was remediated the next day; \S\ref{sec:findings-bimodal}):

\begin{itemize}
\item On the tenant-key access axis (\(\mathtt{GET}\)-by-id with the probing tenant in the query string), tenant-key GET exposure was \(164/164 = 1.000\) (95\% Wilson CI \([0.977, 1.000]\)) and the corresponding in-scope miss rate was \(0/28 = 0.000\). These bulk probes use a tenant-scoped key, which is tenant-wide by design, so this number is not by itself the sub-tenant policy violation. Rather, it shows that once a tenant-authorized caller holds a \(\mathtt{memory\_id}\), the direct access path returns the row. Whether fleet/agent scope is enforced for agent-scoped callers is tested by the focused probe below, which at measurement time showed that sub-tenant scope was not enforced on \(\mathtt{GET}\)-by-id.
\item On the secondary \textbf{discovery} axis (semantic search with the probing fleet in \(\mathtt{fleet\_ids}\)), \textsc{search\_leak\_rate} \(= 72/164 = 0.439\) and \textsc{search\_miss\_rate} \(= 4/28 = 0.143\). The search path \emph{partially} honours the fleet filter --- roughly \(44\%\) of cross-fleet probes still surfaced the row, and a seventh of same-fleet probes missed it under embedding-similarity ranking.
\end{itemize}

A natural objection is that the access leak is merely an artifact of the tenant key carrying no agent identity --- give the gateway an agent identity to filter on and the leak would close. We tested that objection directly with the agent-scoped probe (\S\ref{sec:method}), and at measurement time it did not hold. The gateway \emph{does} resolve the agent identity for a \(\mathtt{kind{=}agent\_key}\) credential (\texttt{/whoami} returns the bound \(\mathtt{agent\_id}\); a tenant key returns \(\mathtt{agent\_id{=}null}\)), yet \(\mathtt{GET}\)-by-id still returned cross-fleet rows: an agent bound to \(\mathtt{fleet\text{-}0}\) at \(\text{trust}=1\) --- below the trust-\(2\) cross-fleet-read rung --- successfully fetched a \(\mathtt{fleet\text{-}1}\) row, while its same-fleet control and the trust-\(2\) credential behaved as expected (Table~\ref{tab:agentscope}). Inspecting the handler confirmed the cause: \(\mathtt{GET\ /memories/\{id\}}\) checked only \(\mathtt{enforce\_readable\_tenant}\) and \(\mathtt{memory.tenant\_id == tenant\_id}\); it never consulted the caller's fleet, agent, the row's visibility scope, or the trust level, so the resolved \(\mathtt{agent\_id}\) was discarded. The access leak was therefore a genuine \emph{enforcement gap} on the \(\mathtt{GET}\) path, not an authentication-model artifact. In the terms of \S\ref{sec:problem}, the handler discharged only the tenant projection of \(s_i\), violating \textsc{Inv-Scope} for fleet- and agent-scoped rows --- the measured leak was exactly the predicted violation. We disclosed the gap; it was remediated server-side during the study (Remediation, \S\ref{sec:findings-bimodal}).

\begin{table}[H]
\centering
\small
\caption{Agent-scoped \(\mathtt{GET}\)-by-id probe, \emph{as measured} (2026-05-30). The gateway resolves each agent's identity (confirmed via \texttt{/whoami}); the documented trust ladder defines cross-fleet read at \(\text{trust}\ge 2\). At that time every probe returned the row (HTTP \(200\)) --- including the trust-\(1\) cross-fleet read the ladder should deny --- because the handler checked only the tenant. \textbf{Since remediated:} a scaled re-probe (2026-05-31, \(96\) probes) denies trust-\(1\) cross-fleet reads (\(0/36\), now HTTP \(404\)) and admits trust-\(2\) (\(36/36\)); see Remediation, \S\ref{sec:findings-bimodal}.}
\label{tab:agentscope}
\begin{tabularx}{\linewidth}{Xll}
\toprule
Caller credential & Target (relation) & Result \\
\midrule
tenant key (\(\mathtt{agent\_id{=}null}\)) & fleet-1 & \(200\) \\
agent, fleet-0, \(\text{trust}=1\) & fleet-1 (cross-fleet) & \(\mathbf{200}\) \emph{(should deny)} \\
agent, fleet-0, \(\text{trust}=2\) & fleet-1 (cross-fleet) & \(200\) (allowed) \\
agent, fleet-0, \(\text{trust}=1\) & fleet-0 (same-fleet) & \(200\) (control) \\
\bottomrule
\end{tabularx}
\end{table}

\begin{figure}[H]
\centering
\includegraphics[width=\linewidth]{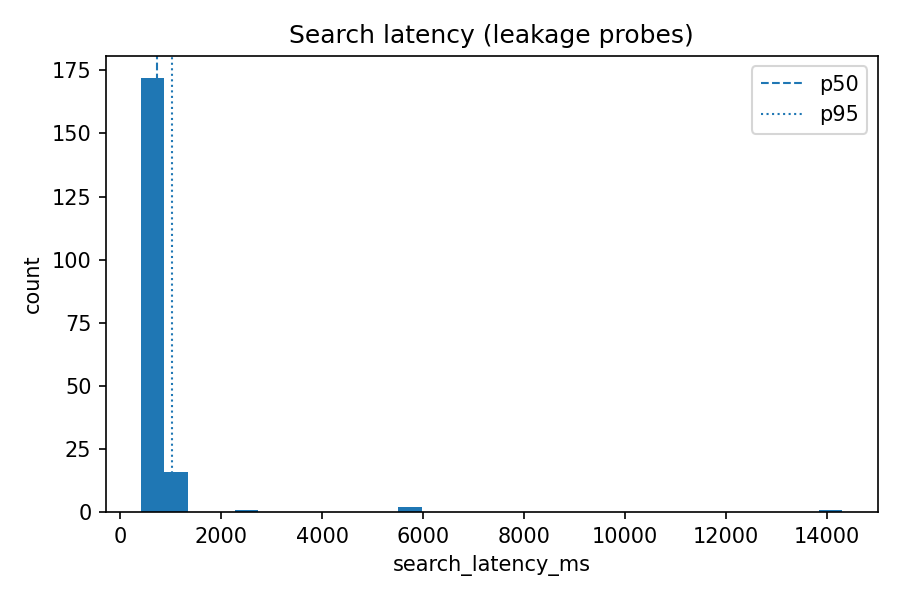}
\caption{Search-latency distribution across the \(192\) leakage probes. Dashed and dotted vertical lines mark the p50 and p95 of the distribution.}
\label{fig:leakage-latency}
\end{figure}

Access latency is tighter than search latency (p50 \(=296\) ms vs \(726\) ms), shown in Fig.~\ref{fig:leakage-latency}, consistent with the access path being a direct primary-key lookup while the search path runs an embedding-similarity scan.

\subsection{Provenance}

We observed complete end-to-end chain reconstruction with sub-second per-hop latency. All \(200\) planned writes landed and all \(50\) depth-four chains reconstructed: completeness, writer-identity accuracy, and depth-fidelity were each \(1.00\), as summarized in Table~\ref{tab:headline}. The harness walked each four-hop derivation chain via \texttt{metadata.derived\_from} at p50 \(=291\) ms and p95 \(=491\) ms per fetch (median sum-per-chain \(\approx 1.2\) s, p99 \(=1.1\) s). In every chain the fetched ancestor's \texttt{agent\_id} matched the planned writer, demonstrating that the server preserves identity through the storage round-trip.

\begin{figure}[H]
\centering
\includegraphics[width=\linewidth]{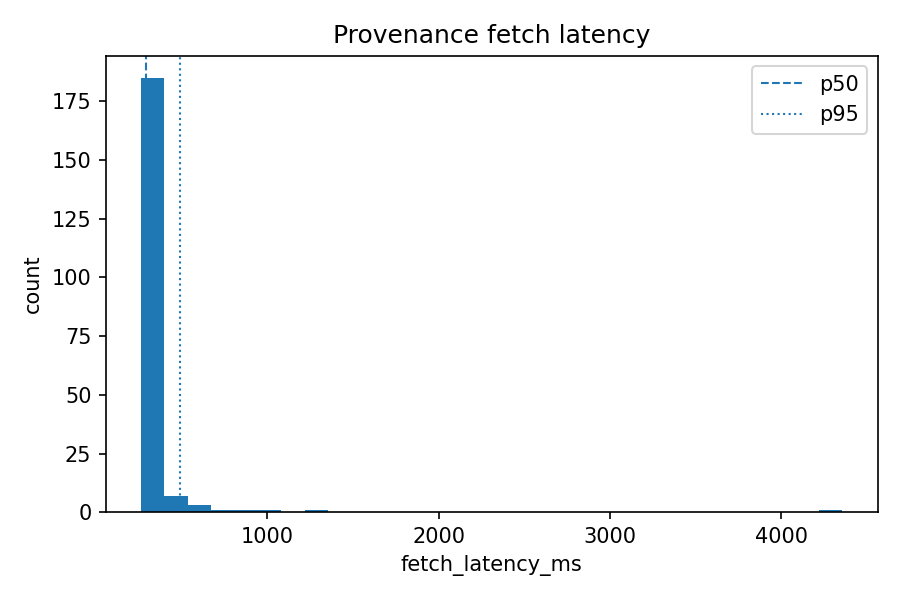}
\caption{Per-hop fetch-latency distribution across all provenance chain walks. The p50 of \(291\) ms and p95 of \(491\) ms support chain reconstruction at interactive latencies even at depth four (p99 \(=1.1\) s).}
\label{fig:prov-latency}
\end{figure}

Reaching complete reconstruction required pacing writes beneath MemClaw's per-tenant write ceiling. The server caps writes at \(10\) per second per tenant (HTTP \(429\), ``\(10\) per \(1\) second''); because provenance writes are sequential within a chain, an unthrottled run whose concurrency bursts past that ceiling aborts the remainder of any chain whose intermediate write is rejected. Throttling the client to \(8\) writes/second --- below the ceiling --- eliminates the rejections without altering the workload, and is what separates the complete reconstruction reported here from earlier runs that lost chains to rate-limited writes. The per-fetch latency distribution is shown in Fig.~\ref{fig:prov-latency}.

\subsection{Propagation}

All \(40\) planned writes landed and all \(200\) downstream visibility probes fired. Fleet-sibling visibility was \(\mathbf{117/120 = 0.975}\) (\(95\%\) Wilson score interval \([0.929, 0.991]\)) across intra-fleet probes, with zero cross-fleet leaks across \(\mathbf{80}\) foreign-fleet probes (\textsc{leak\_rate} \(= 0/80\), \(95\%\) Wilson CI \([0.000, 0.046]\)). The three misses are siblings whose canary never surfaced before the \(30\)-second poll deadline --- a retrieval-recall effect (with many similar facts in the fleet, the target can rank below the top-\(k\)), not a propagation failure.

We measure the write-to-visible window separately, on \(8\) dedicated facts each polled tightly from write completion at \(250\) ms granularity (\S\ref{sec:method}). Under \texttt{write\_mode\,=\,strong} the row is searchable the \emph{instant} the write returns: all \(8\) facts surfaced on the poller's \emph{first} search after the write (\(\mathtt{poll\_count}=1\)), with a median time-to-visible of \(0.83\) s and a max of \(1.97\) s --- itself essentially one search round-trip, indistinguishable from the search-latency floor (leakage search p50 \(\approx 0.73\) s). Strong-mode enrichment is therefore \emph{synchronous}: its cost is paid in the write latency (p50 \(\approx 1.8\) s, \S\ref{sec:results}), and there is no measurable post-write eventual-consistency tail.

\begin{figure}[H]
\centering
\includegraphics[width=\linewidth]{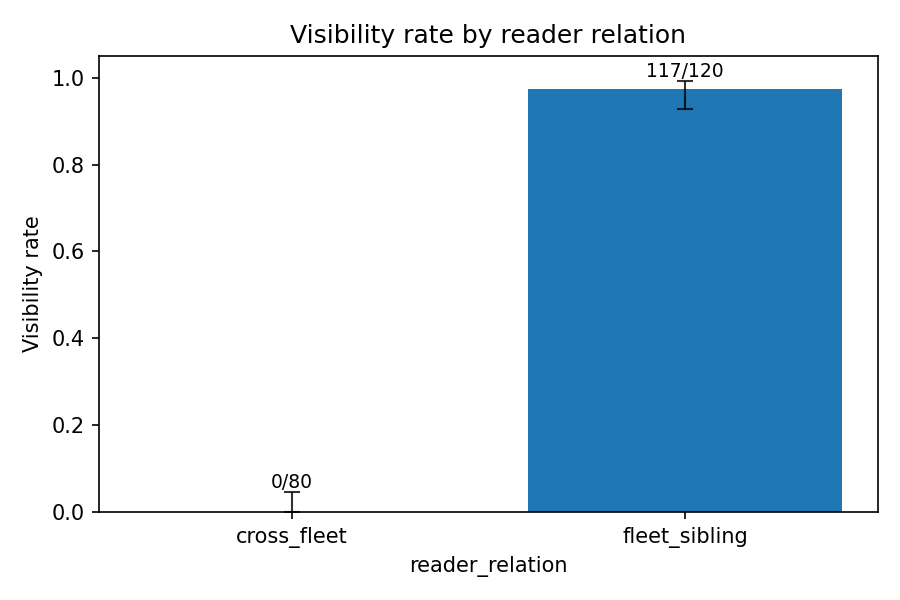}
\caption{Visibility rate by reader relation. Fleet siblings (intra-fleet readers) saw nearly every fact (\(117/120\)); the three misses are writes whose canary ranked below the top-\(k\) under embedding-similarity search (a recall effect, not a propagation failure). Foreign-fleet readers (same tenant, different fleet) saw none (\(0/80\)). Annotations show (visible~/~total) probe counts per axis.}
\label{fig:prop-visibility}
\end{figure}

\begin{figure}[H]
\centering
\includegraphics[width=\linewidth]{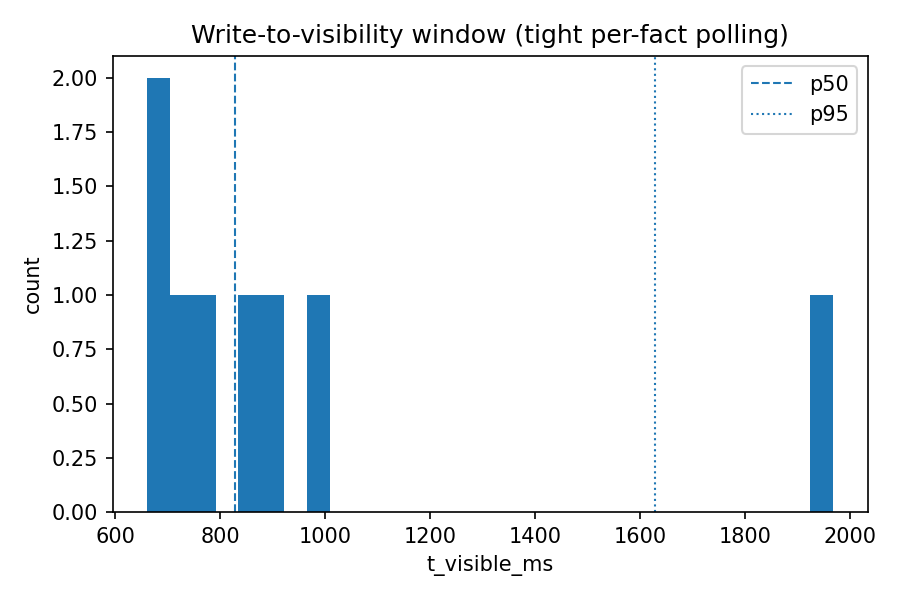}
\caption{Write-to-visible window from the dedicated window phase: \(8\) facts, each polled tightly from write completion at \(250\) ms granularity. Under strong write mode every fact surfaced on the poller's first search (p50 \(=0.83\) s, max \(1.97\) s) --- one search round-trip, i.e.\ effectively immediate. This isolates the server's single-write consistency window from the probe-scheduling inflation a batched schedule exhibits (\S\ref{sec:limitations}).}
\label{fig:prop-tvisible}
\end{figure}

This revises an earlier, naive measurement. Seeding every fact and \emph{then} draining all probes through a concurrency gate makes a probe's first \(\mathtt{/search}\) land long after the row is already searchable; that schedule reported a p50 of tens of seconds, which grows with probe count. It is a probe-scheduling artifact, not server behaviour --- which is precisely why we measure the window with the dedicated tight-polling phase instead (\S\ref{sec:limitations}). The rate phase's fleet-sibling probes still poll \texttt{/search} up to a \(30\)-second deadline (cross-fleet readers issue a single attempt), but only their visibility/leak \emph{outcomes} are used, not their latency. Fig.~\ref{fig:prop-visibility} contrasts the intra-fleet and cross-fleet visibility rates side by side, and Fig.~\ref{fig:prop-tvisible} shows the window-phase time-to-visibility distribution.

\subsection{Contradiction}

Measured correctly, the contradiction primitive \emph{works}. Conditional on both contradictory writes being admitted, supersession was established in \textbf{every} case --- \(90/90 = 1.000\) (\(95\%\) Wilson CI \([0.959, 1.000]\)): a post-commit re-fetch found the newer row carrying \texttt{supersedes\_id} pointing at the older, and the older flipped to \texttt{outdated}, within \(\approx\!6\) s of the write. Overall \textsc{detection\_rate} is \(98/200 = 0.490\); the gap from \(1.0\) is entirely an \emph{admission} effect, not a detector failure (see below).

Two corrections to an earlier, mis-measured version of this experiment underlie the result. First, detection is \emph{post-commit and asynchronous}: \texttt{supersedes\_id} is set after the write response returns, so reading it from the synchronous response (as we first did) reports a spurious \(0/200\). The re-fetch (\S\ref{sec:method}) is what surfaces the true outcome. Second, the structural detector only runs for predicates the server treats as single-valued; our earlier \texttt{deployment\_region} was not one, so the RDF path was silently skipped. Using a single-valued predicate (\texttt{region}, \texttt{runs\_on}, \dots) and re-fetching post-settle fixes both.

The \(0.490\) ceiling is set by write \emph{admission}: \(206/400\) writes were rejected \(409\) by the synchronous dedup gate before reaching the (post-commit) detector, so \(90\) fact-runs had both writes land. The rejections fall roughly evenly across the first and second write of a pair and are dominated by exact-content duplicates --- partly the genuine dedup-pre-empts-contradiction effect of \S\ref{sec:findings} (a contradiction phrased as ``\(X\) is \(A\)'' vs ``\(X\) is \(B\)'' is near-identical text), partly limited prose diversity in our workload generator. We do not cleanly separate the two; the robust claim is the conditional one: \emph{when both writes are admitted, supersession is correct \(90/90\)}. Admission is stochastic --- concurrent writes race at the dedup gate and content is nonce-salted per run --- so counts vary run-to-run around a stable rate: four independent \(N{=}200\) runs gave both-admitted \(90\)--\(91\) and detections \(96\)--\(100\) (\textsc{detection\_rate} \(0.48\)--\(0.50\)).\footnote{The run of record's both-admitted \(=90\) is exactly \(2\times\) the \(N{=}100\) run's \(45\); this is coincidence, not a scaled figure --- sibling runs give \(91\), and all five \(N{=}200\)/\(N{=}100\) contradiction traces are committed under \texttt{traces/} for direct per-write verification.} The overall detection count (\(98\)) slightly exceeds both-admitted (\(90\)) because a fact's sequential and concurrent scenarios share a \texttt{subject\_entity\_id} and predicate, so the single admitted write in one scenario can supersede the admitted write in the other: \(8\) such cross-scenario detections (\(4\) fact pairs) arise where each scenario admitted only one of its two writes. \textsc{stale\_read\_rate} was \(0/200\) (no fact-run returned two active rows), consistent with the older row being marked \texttt{outdated} on resolution; with search visibility now measured at \(\sim\!1\) s (\S\ref{sec:results}), well inside the \(20\) s settle, this signal is no longer materially confounded by search lag, though the GET-by-id re-fetch remains the primary one.

\begin{figure}[H]
\centering
\includegraphics[width=\linewidth]{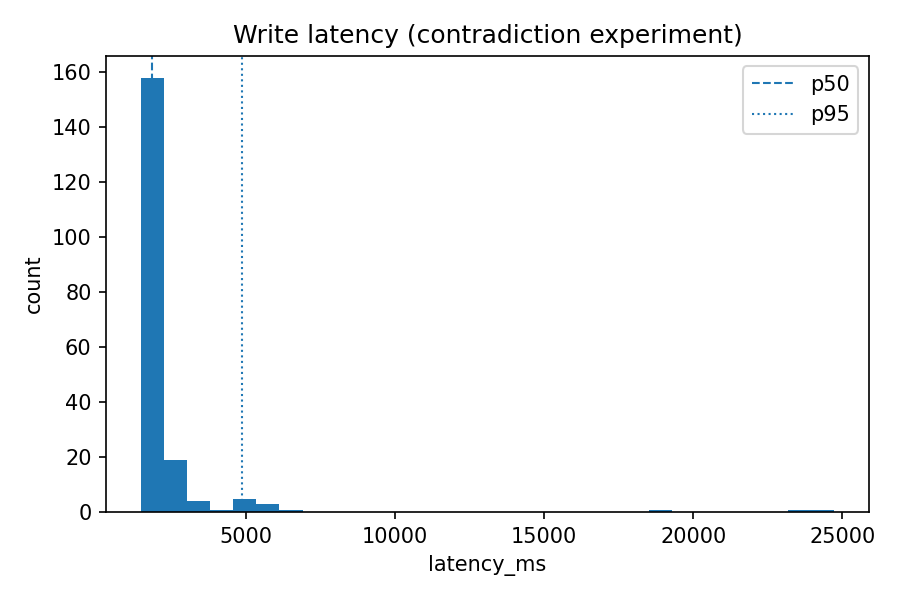}
\caption{Write-latency distribution under \texttt{write\_mode = strong} for the contradiction experiment, in milliseconds (p50 \(1{,}840\), p95 \(4{,}861\); dashed/dotted markers). Contradiction detection runs post-commit and asynchronously, so it contributes nothing to this synchronous write latency: the distribution characterises the steady-state strong-mode write path, with the supersession established afterward (\(\approx\!6\) s, \S\ref{sec:results}).}
\label{fig:contra-write}
\end{figure}

Write-latency p50 of \(1{,}840\) ms and p95 of \(4{,}861\) ms (p99 \(19{,}319\) ms, across the \(194\) writes that completed) reflect the cost of the full enrichment pipeline (\texttt{write\_mode = strong}); the heavier p95/p99 tail relative to the smaller run is queueing when more writes contend for the synchronous pipeline at once. Because contradiction detection is post-commit and asynchronous, it adds nothing to this synchronous write latency. Fig.~\ref{fig:contra-write} shows the distribution.

\section{Discussion and Findings}
\label{sec:findings}

The empirical study surfaces three architecturally significant findings about production memory infrastructure, independent of the headline rate metrics.

\subsection{Scope Enforcement Was Bimodal (GET-Path Gap Remediated)}
\label{sec:findings-bimodal}

The leakage experiment surfaces a structural asymmetry in how scope is enforced. Tenant boundaries are enforced uniformly: cross-tenant requests are rejected on both \(\mathtt{GET}\)-by-id and \(\mathtt{POST}\)-search paths (the leakage rate measurements above were collected within a single tenant for that reason --- cross-tenant probes do not reach the read handlers at all). \emph{Sub-tenant} scope --- fleet and agent --- was, at measurement time, enforced only on the search path (and there only partially); the \(\mathtt{GET}\)-by-id path is addressed by the remediation below.

At measurement time, direct \(\mathtt{GET}\)-by-id returned a row to any tenant-authorized caller who presented the correct \(\mathtt{tenant\_id}\), regardless of the originating fleet or writer. Crucially, this held even when the caller presented an agent-scoped credential whose identity the gateway resolves (\S\ref{sec:results}, Table~\ref{tab:agentscope}): the \(\mathtt{get\_memory}\) handler enforced only the tenant predicate (\(\mathtt{enforce\_readable\_tenant}\) plus \(\mathtt{memory.tenant\_id == tenant\_id}\)) and never consulted the caller's fleet/agent, the row's visibility scope, or the trust level --- so the agent identity, present in-request as \(\mathtt{X\text{-}Agent\text{-}ID}\), was simply discarded. Search, by contrast, accepts \(\mathtt{fleet\_ids}\) in the request body and applies it as a filter, but the filter coexists with embedding-similarity ranking: \(43.9\%\) (\(72/164\)) of cross-fleet probes still surfaced the row.

We initially conjectured this asymmetry was a property of the authentication model rather than a defect --- that agent-scoped keys would let the gateway resolve a requesting agent's fleet and apply scope to \(\mathtt{GET}\) as it does to search. The probe in \S\ref{sec:results} refuted that conjecture: the identity \emph{was} resolved and then ignored by the handler, so a trust-\(1\) agent read a cross-fleet row the trust ladder should deny. It was therefore a genuine enforcement gap --- a violation of \textsc{Inv-Scope} (\S\ref{sec:problem}) --- not a credential limitation, and while it stood, the propagation of \(\mathtt{memory\_id}\) values across fleet boundaries was itself the de-facto capability boundary.

\paragraph{Remediation.} We disclosed the gap; it was fixed server-side during the study. A scaled re-probe on 2026-05-31 --- \(96\) \(\mathtt{GET}\)-by-id probes across four fleets and both trust levels, with each agent using its own agent-scoped key, committed under \texttt{traces/} --- confirms the \(\mathtt{get\_memory}\) handler now applies exactly the fleet/agent/scope filter the search path already carried: trust-\(1\) cross-fleet reads are denied (\(0/36\), 95\% Wilson \([0.000, 0.096]\), returned as \(\mathtt{404}\) so the row's existence is not disclosed), trust-\(2\) cross-fleet reads are admitted by the ladder (\(36/36\)), and same-fleet controls pass (\(12/12\)); total leaks \(0/96\). The bulk \(164/164\) access rate (\S\ref{sec:results}) was collected with a \emph{tenant}-scoped key, which is tenant-wide by design and never subject to sub-tenant scope; with agent-scoped reads now enforced, the credential-independence we originally observed no longer holds. We name the resulting asymmetry plainly: the bulk \(164/164\) (tenant key, 2026-05-30) and the \(0/36\) remediation re-probe (2026-05-31) are both committed traces (\texttt{traces/}), but the agent-scoped \emph{open} state --- the surprising part --- is no longer reproducible against the live service, so it rests on the dated probe (Table~\ref{tab:agentscope}) and the \(\mathtt{get\_memory}\) handler reading rather than a replayable trace. The verifiable artifacts thus attest the tenant-key behaviour and the fix; the transient agent-scoped gap is dated and inspected, not replayable. The episode is itself a result: the formalism flagged a concrete \textsc{Inv-Scope} violation, the harness localised it to one handler, and the remediation was precisely the predicate the search path already carried --- evidence that the failure-mode taxonomy points at fixable, well-localised defects rather than diffuse ones.

\subsection{Deduplication Starves Contradiction Detection}
\label{sec:findings-dedup}

The contradiction detector itself is sound: conditional on both writes being admitted, it superseded correctly in every case (\(90/90\), \S\ref{sec:results}). The architectural issue is that two mechanisms on the write path interact badly. A semantic near-duplicate detector evaluates an incoming write synchronously, \emph{pre-commit}, returning \(409\) when embedding similarity exceeds a threshold. The contradiction detector evaluates RDF-triple compatibility and sets \texttt{supersedes\_id} --- but it runs \emph{post-commit and asynchronously}. So the synchronous gate can reject a write before the asynchronous detector ever sees it. The pathological case is intrinsic: a contradiction phrased naturally (``\(X\) is \(A\)'' then ``\(X\) is \(B\)'') is near-identical text, so the very writes the detector exists to resolve are the ones most likely to be 409'd at the gate. (This is also why the detector reads as a ``null'' under naive measurement --- it never receives the input, and \texttt{supersedes\_id} is async besides.)

We argue this ordering is incorrect: a contradiction is precisely a near-duplicate with a different assertive object, and conflating the two collapses two distinct architectural primitives into one error code. We propose either (a) reordering the pipeline so that structural contradiction detection runs first when a write carries an RDF triple, or (b) widening the near-duplicate threshold for writes that carry an RDF triple, since the structural signal is the more discriminating one.

\subsection{Cross-Agent Propagation Is Privilege-Gated}

The \texttt{POST /memories/redistribute} endpoint requires the requesting agent to hold \(\text{trust\_level} \ge 3\), placing cross-agent memory propagation behind an administrative gate. From the perspective of the architecture in \S5, this conflates two policy concerns: \emph{can this agent propagate fleet-shared knowledge to a sibling within its own fleet?} and \emph{can this agent administratively re-home memories arbitrarily?} The first is a normal operation in any non-trivial multi-agent workflow; the second is a privileged migration.

We argue that fleet-scoped propagation between sibling agents inside a fleet should be available at the same trust level as fleet-scoped writes, and that the higher trust gate should apply only to cross-fleet or cross-scope re-homing. The current production behaviour is consistent with treating all redistributes as administrative migrations; under a governed-memory architecture the propagation primitive should expose a finer-grained policy surface.

\subsection{The Consistency Cost Is Paid at Write Time, Not Read Time}

Measured with tight per-fact polling \emph{at low contention} (the window phase polls one fact at a time; \S\ref{sec:results}), the write-to-visible window under \texttt{write\_mode\,=\,strong} is effectively zero: a fleet sibling's first search after the write already returns the row (p50 \(\approx 0.8\) s --- one search round-trip). Strong-mode enrichment is \emph{synchronous}, so the consistency cost is paid in the write latency, not as a post-write read-visibility lag --- but that cost is load-dependent: at the sustained write concurrency of the contradiction run the write-latency p99 reaches \(19\) s (\S\ref{sec:results}), against a p50 of \(\approx\!1.8\) s. The window phase, deliberately uncontended, measures the visibility floor; the write-latency tail is the price the same consistency guarantee charges under load.

This matters methodologically as much as architecturally. A naive measurement that seeds every write and \emph{then} drains read probes reports a window of tens of seconds that \emph{grows with probe count} --- the entire figure is harness scheduling, not server behaviour, because a drained probe's first look lands long after the row is already searchable. (An earlier version of this work reported exactly such a figure.) Quantifying a service's true consistency window therefore requires polling each write immediately and continuously, decoupled from the rate workload; otherwise one measures one's own client. The architectural takeaway is robust to that subtlety: the placement of enrichment on the write path versus the read path is a first-order design choice, and a service that enriches synchronously (as MemClaw does under strong mode) buys immediate read consistency at the price of write latency --- the budget a multi-agent orchestrator must plan against is the write, not a visibility tail.

\section{Limitations}
\label{sec:limitations}

Several limitations bound the scope of the empirical claims in \S\ref{sec:results}.

\paragraph{Self-evaluation.}
ArgusFleet and MemClaw are built by the same group; this is an evaluation of our own production service, not an independent audit. We mitigate the obvious bias by reporting the unflattering results prominently --- the \(\mathtt{GET}\)-by-id enforcement gap (disclosed, since remediated) and the dedup/contradiction pipeline-ordering issue --- and by deriving every figure from a committed JSONL trace. Reporting a gap in one's own service and then its fix is an obvious place for bias to hide; we counter it by committing the re-probe trace, not just asserting the remediation. An independent replication, which the open harness enables against any wire-compatible service, would carry more weight than our own numbers.

\paragraph{Agent-scope probe is focused, not exhaustive.}
The bulk leakage rates were collected with a tenant-scoped key (tenant-wide by design); the \emph{sub-tenant} question was settled with agent-scoped credentials. The original observation was a small focused probe (Table~\ref{tab:agentscope}); the post-remediation re-probe is larger (\(96\) \(\mathtt{GET}\)-by-id probes across four fleets and both trust levels, committed under \texttt{traces/}) and finds the gap closed (\(0/36\) trust-\(1\) cross-fleet, \S\ref{sec:findings-bimodal}). Two caveats follow from the timeline: the open state was observed at one point and is no longer reproducible against the live service, so the as-measured gap rests on the dated observation plus the \(\mathtt{get\_memory}\) handler reading, while the remediation rests on the committed re-probe; and a fuller adversarial matrix (many agents, scopes, visibility levels) would harden both. Cross-\emph{tenant} isolation remains untested for the separate reason below.

\paragraph{Staleness-after-supersession is only partially measured.}
The contradiction experiment establishes the \emph{resolution} side of staleness: when both writes are admitted the older row flips to \texttt{outdated} and the newer carries \texttt{supersedes\_id} (\S\ref{sec:results}, \(90/90\)), and \(\textsc{stale\_read\_rate} = 0/200\) found no fact-run serving two simultaneously-active rows. But the stale-read check rides on semantic search, whose enrichment window we now measure directly at \(\sim\!1\) s (\S\ref{sec:results}) --- well within the \(20\)-second settle --- so a \(0\) there reflects correct resolution rather than search lag. (An earlier draft, which inferred the window from the batched probe schedule rather than measuring it, could not separate the two.) The GET-by-id re-fetch remains the primary signal; the \texttt{active}-search exclusion now corroborates it instead of being confounded by lag.

\paragraph{Single tenant.}
All measurements were collected from a single MemClaw tenant, freshly provisioned. The cross-\emph{tenant} leakage probe surface, which would exercise the strongest governance claim, is not measurable in this configuration. The harness supports multi-tenant configuration via the \texttt{MEMCLAW\_TENANT\_ID} list and we expect cross-tenant probes to be uniformly rejected by the gateway's tenant enforcement (\S\ref{sec:findings-bimodal}); confirming that empirically remains future work.

\paragraph{Workload scale.}
We report \(200\) trials per experiment. The \(\textsc{leak\_rate}\) Wilson interval at \(164/164\) is \([0.977, 1.000]\); the \(\textsc{cross-fleet leak rate}\) at \(0/80\) in propagation is bounded above by approximately \(0.046\). Stronger claims would benefit from a further order of magnitude more probes and probes constructed adversarially rather than via deterministic plan generation.

\paragraph{Time-to-visibility requires decoupled measurement.}
Time-to-visibility cannot be read off the rate-phase probes: seeding every write and then draining all probes through a concurrency gate makes each probe's first \(\mathtt{/search}\) land after the row is already searchable, so the recorded latency measures harness scheduling and \emph{grows with probe count} (an earlier version of this work reported a p50 of tens of seconds for exactly this reason). We therefore measure the window on a small set of facts polled immediately and continuously from write completion (\S\ref{sec:method}), one at a time so the polls stay under the \(\mathtt{/search}\) budget; under strong mode this yields a p50 of \(\sim\!0.8\) s (\S\ref{sec:results}). The residual limitation is granularity and scale: the window is measured at \(250\) ms resolution on \(8\) facts, so it resolves ``effectively immediate'' but not a sub-\(250\) ms value, and a larger, multi-region sweep would sharpen the tail and test whether the synchronous-enrichment behaviour holds under sustained load.

\paragraph{No comparand.}
We do not compare against a baseline (e.g., a long-context-only configuration, or an alternative memory implementation). The architectural claims of \S5 stand on the failure-mode taxonomy and the implementation, but the empirical claims would be sharper with a comparison condition.

\section{Conclusion}

Taken together, the paper delivers the three contributions introduced in \S1.1: a formal framing of fleet memory as governed operational state, an implementation and evaluation harness for exercising the proposed primitives, and a live-service measurement showing both where those primitives hold and where production enforcement and pipeline ordering can fail.

We introduced a systems architecture for governed shared memory in multi-agent LLM environments, instantiated it in MemClaw, and evaluated the resulting primitives using ArgusFleet, a reproducible harness that exercises the four governance failure modes of \S\ref{sec:failures} as four experiments against a live memory service, reporting five metric axes: governance correctness, contradiction handling, provenance reconstruction, and cross-agent visibility --- one per failure mode --- plus \emph{retrieval recall} as an orthogonal availability control sharing its probe set with the leakage experiment.

The empirical study shows that provenance chains reconstruct completely (\(50/50\)) at sub-second per-hop latency (p50 \(\approx 291\) ms at depth four) once writes are paced below the server's rate ceiling, that cross-fleet propagation is correct on \(117/120\) intra-fleet probes (\(0.975\)) with zero leaks across \(80\) foreign-fleet probes, and that --- measured by tight per-fact polling at low load --- the write-to-visible window under strong write mode is effectively immediate (p50 \(\approx 0.8\) s, one search round-trip), with the consistency cost paid in write latency (which itself develops a heavy tail, p99 \(\approx 19\) s, under sustained write contention) rather than as a read-visibility lag. It also surfaces three architectural observations: scope enforcement was bimodal at measurement time --- tenant-level isolation enforced on every path, sub-tenant scope enforced partially on the search path and not at all on direct \(\mathtt{GET}\)-by-id (the read handler discarded even a resolved agent identity); we disclosed this and it was remediated server-side during the study, with a committed re-probe confirming agent-scoped \(\mathtt{GET}\)-by-id now enforces fleet scope and the trust ladder; contradiction supersession is correct when both writes are admitted (\(90/90\)), but a synchronous dedup gate pre-empts near-identically phrased contradictory writes before the asynchronous detector runs; and cross-agent memory re-homing is privilege-gated in a way that conflates fleet-internal propagation with administrative migration.

We argued throughout that AI memory is evolving from a context-window problem into a distributed systems problem requiring governance, synchronization, temporal correctness, provenance, and scoped access control. The empirical results support this framing: in production memory infrastructure, the primitives that matter for correctness are precisely the ones inherited from distributed systems --- consistency windows, identity preservation through round-trips, ordering of pipeline stages, and the privilege model attached to propagation. As AI systems continue shifting from isolated assistants toward coordinated fleets of agents, governed shared memory becomes a foundational infrastructure layer whose architectural choices have measurable consequences for the workflows built on top.

\bibliographystyle{plainnat}
\bibliography{references}

\end{document}